# An Image-based Generator Architecture for Synthetic Image Refinement

Alex Nasser

**Abstract**: *Proposed are alternative generator architectures for Boundary Equilibrium Generative Adversarial Networks, motivated by Learning from Simulated and Unsupervised Images through Adversarial Training. It disentangles the need for a noise-based latent space. The generator will operate mainly as a refiner network to gain a photo-realistic presentation of the given synthetic images. It also attempts to resolve the latent space's poorly understood properties by eliminating the need for noise injection and replacing it with an image-based concept. The new flexible and simple generator architecture will also give the power to control the trade-off between restrictive refinement and expressiveness ability. Contrary to other available methods, this architecture will not require a paired or unpaired dataset of real and synthetic images for the training phase. Only a relatively small set of real images would suffice.*

## 1. Introduction

Advancement in 3D modelling tools and the ability to render highly realistic faces for immersive purposes in computer games and graphics have always been costly and challenging. Insufficient graphical processing power in rendering these models to simulate all given real-world parameters could be one of the main factors contributing to facial rendering artifacts and shortcomings [1]. The end-product requires encapsulating accuracy, not just in terms of illumination and reflectance of facial details, but also taking skin's spatial correlation like facial wrinkles into account [12]. Thus, making the use of synthetic images, produced through recent advancements in Deep Learning methods, a more attainable option.

Furthermore, in the field of Machine learning, there is an increasing need for large labelled training datasets. Curating such datasets and labelling them requires time and workforce. Therefore, synthetic images seem to be a valid replacement for real-life data. The refiner model can reduce the disparity between synthetic and real images, creating the need to modify these images to their photorealistic presentation [11].

Generative Adversarial Networks (GANs) can produce photorealistic facial rendering, and since its introduction [2], there has been a variety of papers related to numerous architectures, approaches of training these models and their relevant loss functions that improved the quality and stability of these networks. Novel training methods now gives us the ability to create high resolution photorealistic images [3][4] with some networks focusing on stability during training with changes to the loss functions and architecture [5][6][7]. Nevertheless, the absence of an easy architecture or training method to achieve full control of the generator's latent space has proven difficult [8][9][10].

Inspired by Learning from Simulated and Unsupervised Images through Adversarial

Training (SimGAN) [11] and derived from Boundary Equilibrium Generative Adversarial Networks (BEGANs) [7], proposed here, are new generator architectures that uncover image-based solutions for latent space. This is achieved by replacing the traditional random noise injection with a series of downsized versions of the training sample. The level of downsizing and the number of inputs dictates the restriction given to the network for generating a photorealistic presentation of given images. The network follows the BEGAN's loss function with the addition of a content loss for its generator.

## 2. Proposed Method:

### 2.1. Model Architecture:

Similar to Energy-based Generative Adversarial Network (EBGAN) and BEGAN, the discriminator uses an auto-encoder architecture, which focuses on matching auto-encoder loss distribution and an equilibrium term that balances the generator and the discriminator. The Nearest Neighbour is used for both upsampling and downsampling throughout the discriminator and generator architecture.

The generator architecture, in other words, the refiner, is designed around the concept that quality reduction of input images should ease key facial feature extraction and facilitate image refinement for the model. Refinement based on reconstruction on lost features, compared to a noise-based latent space generative model, will significantly reduce training time. Following the proposed idea of different sizes of input, in an aim to unravel the latent space for the purpose of refinements, the level of quality reduction of input will, in turn, define the expressiveness of the model. This is gained by the number of different levels of downscaled images provided as inputs, similar to skip connections that prevent progressive information reduction. The unsupervised nature of this model and its reliance on only real-life photos can extend the benefits of the refinement to even tools like 3D avatar creators for websites. Tools that typically lack complicated rendering engines and only present key facial features of a face model. A trained model would refine even those images whilst it is only ever trained on real-life photos.

Similar to the BEGAN's discriminator architecture, as shown in Figure 1, the network uses a simple deep autoencoder design using 3x3 kernels, alongside Exponentiation Linear Units (ELUs) as the activation layer. With the repetition of these layers in each block. This is followed by the Nearest neighbour resizing layer, at the end of each block, to upsample or downsamples the images based on which side of the autoencoder they are. The size of the filter used in the convolution layers, which vary based on the initial filter size set, as described in the original paper, will result in more photorealistic presentations. The highest initial filter size used for the model, as depicted in Figure 1, was 64. The encoder will increase the filters after each block to extract more facial features. The decoder will follow a constant filter size with a fully connecter layer connecting the encoder and the decoder.

Figure 2 shows three different versions of the proposed Generator architecture. It is almost identical to the design of the decoder part of the discriminator with the same number of constant filters across all layers. The crucial difference between this generator and most traditional GAN is using downscaled training images as the initial input. After each resizing, there is an opportunity to inject a higher quality version of the original downscaled images into the network. In this paper, model (A) is the most expressive and (C) the most restrictive refiner.

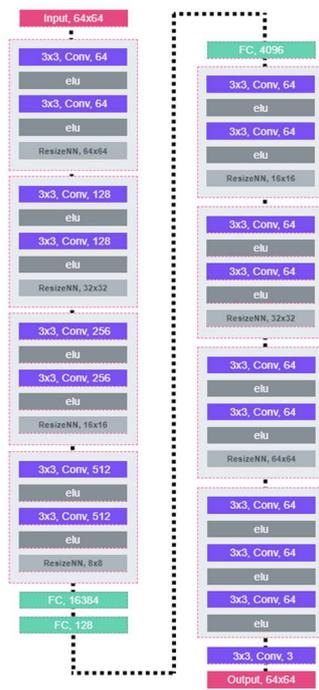

Figure 1: Discriminator Architecture, which follows a simple autoencoder architecture

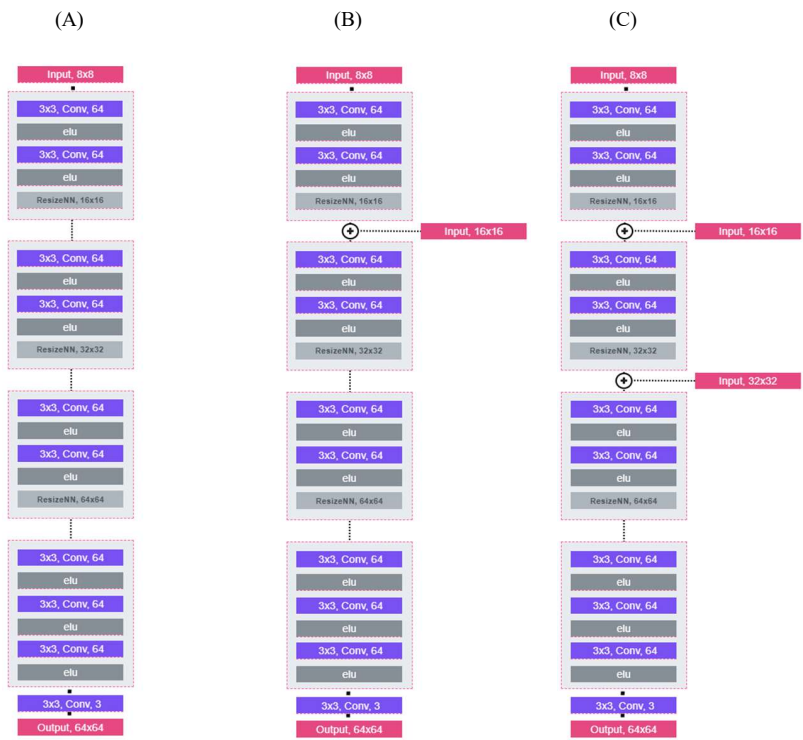

Figure 2: Three different architectures proposed for the generator

## 2.2. Loss Function:

Similar to BEGAN, the network uses an L1-norm for the adversarial loss on pixel-level, on the discriminator's output, $\mathcal{L}_{GAN}$.

$$\mathcal{L}_{GAN}(v) = |v - D(v)|$$

(1)

As shown in the BEGAN research paper, the discriminator loss, $\mathcal{L}_D$, and the generator loss, $\mathcal{L}_G$, are calculated in each training step as shown in equation 2. BEGAN network favours real image distribution. To balance the generator and discriminator loss it also introduces an equilibrium parameter, $k_t$. The discriminator's autoencoder will train and detect anomalies, whereas the generator will train to reduce the detected anomalies by the discriminator.

$$\begin{cases} \mathcal{L}_D = \mathcal{L}_{GAN}(x) - k_t \cdot \mathcal{L}_{GAN}(G(z)) \\ \mathcal{L}_G = \mathcal{L}_{GAN}(G(z)) \\ k_{t+1} = k_t + \lambda_k \left( \gamma \mathcal{L}_{GAN}(x) - \mathcal{L}_{GAN}(G(z)) \right) \end{cases}$$

(2)

In this paper, based on the new architecture proposed, the network's generator loss function is tailored for image refinement through reconstruction. An additional L1-norm reconstruction loss, $\mathcal{L}_{RCN}$, for lower resolution inputs, $v_{LR}$, which calculate the loss between the output of the generator on lower resolution inputs, against the higher resolution versions of it, $v_{LR}$.

$$\mathcal{L}_{RCN}(v_{LR}) = |v_{HR} - G(v_{LR})|$$

(3)

The addition of reconstruction loss, is introduced to the network as a weighted loss to complement the adversarial loss mentioned previously, $\mathcal{L}_{GAN}$. The weight, $\lambda_r$, can explicitly be used to inform the network how much pixel information from low resolution images, should be retrained during refinement.

$$\begin{cases} \mathcal{L}_D = \mathcal{L}_{GAN}(x) - k_t \cdot \mathcal{L}_{GAN}(G(z)) \\ \mathcal{L}_G = ((1 - \lambda_r) \mathcal{L}_{GAN}(G(z))) + \lambda_r \mathcal{L}_{RCN}(z) \\ k_{t+1} = k_t + \lambda_k \left( \gamma \mathcal{L}_{GAN}(x) - \mathcal{L}_{GAN}(G(z)) \right) \end{cases}$$

(4)

As you can see in equation (4), the generator loss has changed to fit the new image-based architecture.

## 3. Experiments:

Here follows some examples of generated faces with the current models. The proposed models were trained on a dataset of 20k front-facing, aligned and centred portraits and validated on rendered 3D models. The decision to limit the images on front-facing photos was to constrain the experiment to pixel refinement.

They were trained using Adam with a learning rate of 0.001. Different versions of the model for generating different resolutions have been used, with 8x8 being the lowest resolution input. Depending on the output resolution, the number of layers was reduced or increased accordingly. Due to hardware limitations, the highest resolution used was 64x64. The 64x64 synthetic refiner models were trained on an Nvidia 1080ti which training of each model, with a batch size of 25, took approximately 48 hours. By mainly focusing on visual analysis, the purpose of the experiment was to evaluate different outputs of each model based on two criteria, photorealistic quality and similarity to the input image.

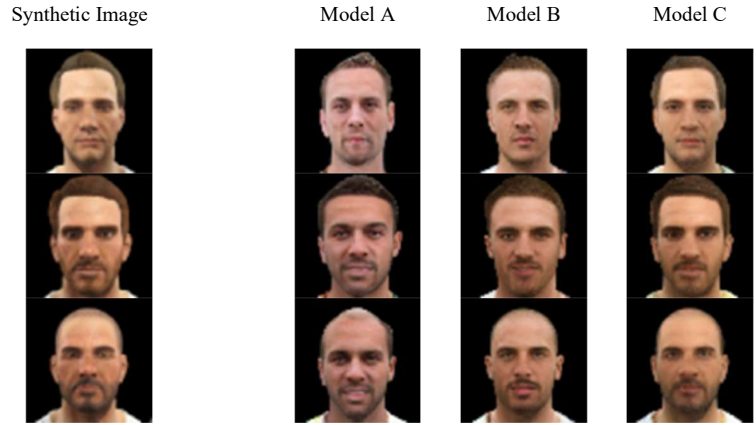

Figure 3: Results from three different architectures based on given 64x64 synthetic images taken from the video game Fallout 4, which are downscaled accordingly, depending on the model's inputs.

In figure 3, you can see the results of different models on three different synthetic images. A survey analysis approach for evaluating the photorealism of the images has been used here which 100 Amazon Mechanical Turk workers ranked the outputs on each model based on realism. The results are shown in figure 4. Based on the percentage shown, Model A produces the most realistic images, likely due to the level of modification freedom given to the network and Model C the least realistic images according to the survey.

| Synthetic Image | Model A | Model B | Model C | Ranked |
|---|---|---|---|---|
| | **58.5** | 36.6 | 4.9 | First |
| | 26.8 | **56.1** | 17.1 | Second |
| | 14.7 | 7.3 | **78** | Third |
| | **50** | 43.9 | 6.1 | First |
| | 18.3 | **50** | 31.7 | Second |
| | 31.7 | 6.1 | **62.2** | Third |
| | **87.8** | 9.8 | 2.4 | First |
| | 4.9 | **73.1** | 22 | Second |
| | 7.3 | 17.1 | **75.6** | Third |

Figure 4: Amazon Mechanical Turk Worker Survey analysis result, ranking models, based on photorealism. (All the values shown are percentage values)

Taking a more quantitative approach towards comparing the similarities between a 3D model and the generated faces, Betaface API, a facial recognition web service that supports face comparisons, has been used. The comparison works based on extracted attributes and facial features. It is worth pointing out that using facial recognition for comparison is not the ideal system. Even though facial recognition and perceived facial similarity are related, they are still considered distinct tasks [14]. As the experiment results show in Figure 5, Model C achieved the highest percentage of similarity.

| Synthetic Image | Model A | Model B | Model C |
|---|---|---|---|
| - | 70% | 69% | **82%** |
| - | 65% | 76% | **90%** |
| - | 69% | 81% | **90%** |

Figure 5: Facial feature comparison of synthetic images and generated images through facial recognition using the BetaFace API web service

The idea behind these experiments was to achieve a visual estimation of whether basic traits of synthetic images can be interpreted by the network to guide the model to generate a refined version. The results as shown, illustrates the expected outcome within the spectrum of confinement and creative freedom, with Model B presenting an acceptable balance between realism and similarity.

4. **Conclusion:**

As seen in the experiment section, work still needs to be done to improve the likeness between the synthetic and refined images without compromising the freedom given to the network. As observed, this freedom is crucial for creating photorealistic presentations through extensive refinement. One method could be to combine the input described in Model A with facial landmarks input as an overlay in the downsampled image and check the results of the refiner against the high-resolution images and the facial landmarks of the generated images against the facial landmarks of the high-resolution images. This should hopefully improve the likeness attributes of Model A.

Another unanswered question is whether different levels of inputs are making any difference in Model B and Model C and whether the network could produce the same results with a simpler architecture. What is clear is the simplicity of training and effectiveness of directing the latent space and providing it with the basic traits of the face without the need for a noise input.

A novel input method was proposed to generate refined images by increasing the resolution that follows the distribution of photo-realistic faces without requiring any pairing or need of synthetic images for training. The method provides a simple solution to create photo-realistic avatars, beneficial for games. A simple pixelated sketch of a face would result in a variety of outputs that can be used to generate game characters or facial textures for 3D images. But there is clearly still significant room for improvement.